\pdfoutput=1

\documentclass[11pt]{article}

\usepackage[preprint]{acl}

\usepackage{times}
\usepackage{latexsym}
\usepackage{bbm}
\usepackage{amsmath}
\usepackage{url}
\usepackage{graphicx}
\usepackage{caption}
\usepackage{subcaption}
\usepackage{multirow}
\usepackage{booktabs}
\usepackage{arydshln}
\usepackage{colortbl}

\usepackage[T1]{fontenc}

\usepackage[utf8]{inputenc}

\usepackage{microtype}

\usepackage{inconsolata}

\title{Improving Weak-to-Strong Generalization with Reliability-Aware Alignment}

\author{Yue Guo  \quad Yi Yang\\
        The Hong Kong University of Science and Technology \\
\texttt{yguoar@connect.ust.hk} \quad \texttt{imyiyang@ust.hk}}

\begin{document}
\maketitle

\begin{abstract}

Large language models (LLMs) are now rapidly advancing and surpassing human abilities on many natural language tasks. However, aligning these super-human LLMs with human knowledge remains challenging because the supervision signals from human annotators may be wrong. This issue, known as the "super-alignment" problem, requires enhancing weak-to-strong generalization, where a strong LLM must generalize from imperfect supervision provided by a weaker source. To address this issue, we propose an approach to improve weak-to-strong generalization by involving the reliability of weak supervision signals in the alignment process. In our method, we query the weak supervisor for multiple answers, estimate the answer reliability, and enhance the alignment process by filtering out uncertain data or re-weighting reliable data. Experiments on four datasets demonstrate that our methods effectively identify the quality of weak labels and significantly enhance weak-to-strong generalization. Our work presents effective techniques for error-robust model alignment, reducing error propagation from noisy supervision and enhancing the accuracy and reliability of LLMs. Codes are publicly available\footnote{\url{https://github.com/Irenehere/ReliableAlignment}}.



\end{abstract}

\section{Introduction}

The large language models (LLMs) are now evolving at an unprecedented pace \cite{DBLP:journals/corr/abs-2303-08774,DBLP:journals/corr/abs-2305-10403,llama3modelcard}. The most advanced LLMs are approaching, and even surpassing, human capabilities in many tasks, such as reading comprehension \cite{DBLP:conf/emnlp/RajpurkarZLL16}, paraphrase identification \cite{DBLP:conf/acl-iwp/DolanB05} and others \cite{DBLP:conf/emnlp/WangSMHLB18}. 
However, training these advanced LLMs usually relies on human supervision or feedback, making aligning super-human models with human knowledge challenging. This challenge is partly because human annotators can produce noisy supervision signals, which can lead to mistakes during the alignment process.

Previous work has identified this challenge and introduced the concept of "super-alignment" \cite{DBLP:journals/corr/abs-2312-09390}. Super-alignment tackles the issue of aligning a super-human model with weaker human supervisors. More generally, the super-alignment is a case of weak-to-strong generalization, which requires a stronger language model to generalize from the noisy supervision signal provided by a weaker supervisor (model or human). The critical challenge in weak-to-strong generalization lies in the unreliability of the weak supervision signals and the inaccessibility of the ground truth.

\begin{figure}
\centering
\includegraphics[width=\linewidth]{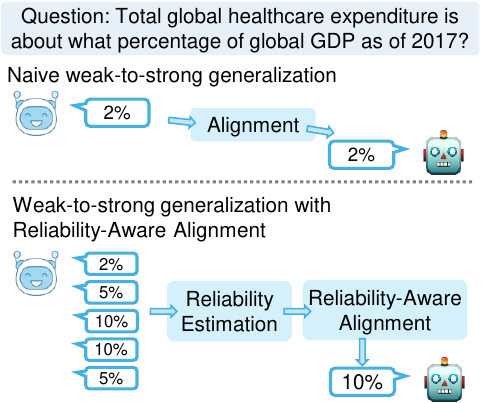}
\caption{Comparison of naive and reliability-enhanced weak-to-strong alignment approaches. 
 The naive weak-to-strong alignment method trains the strong model using the weak labels. Our improved method incorporates reliability estimation on the multiple answers and enhances the alignment process by considering the label reliability, leading to a more accurate response.}
\label{Figure1}
\vspace{-0.2cm}
\end{figure}

\begin{figure*}
\centering
\includegraphics[width=\linewidth]{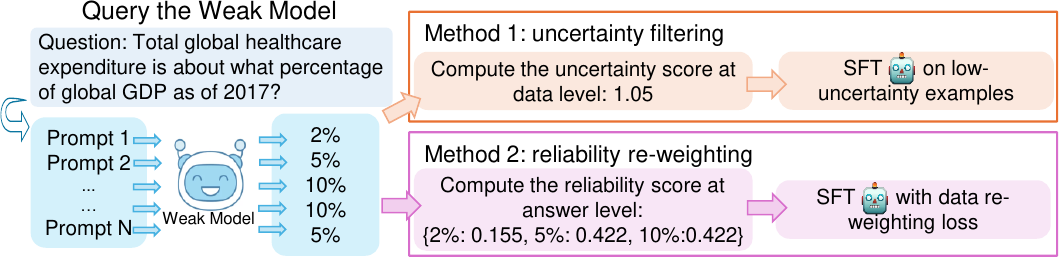}
\caption{The method for enhancing the reliability of the weak-to-strong model alignment. First, we query the weak model for multiple weak labels. Then, we supervised fine-tuning (SFT) the strong model with uncertainty filtering and reliability re-weighting methods. }
\label{Figure2}
\vspace{-0.2cm}
\end{figure*}

To address this challenge, we propose an unsupervised method to enhance weak-to-strong generalization by identifying the reliability of weak supervision signals, thereby improving alignment quality. Figure \ref{Figure1} compares the naive and reliability-enhanced weak-to-strong alignment approaches. In naive weak-to-strong alignment, the strong model learns directly from the weak supervisor's answers (weak labels), making it prone to inherit errors. For more robust alignment results, we propose to query the weak supervisor multiple times and estimate the reliability of the answers using uncertainty and probability-based metrics. Then, the reliability-aware alignment methods using these metrics are applied, leading to better performance for the strong model.

More specifically, our methods for enhancing the reliability of the weak-to-strong model alignment are shown in Figure \ref{Figure2}. First, we augment the original question into a set of variations using simple rules and use these variations to query a set of answers from the weak supervisor. Then, we propose two methods for reliability-aware alignment. In the first uncertainty filtering method, we estimate the reliability of the data using an entropy-based uncertainty metric, then filter out uncertain instances and use the low-uncertainty data for alignment. In the second reliability re-weighting method, we estimate the reliability of the answers using a probability-based reliability metric and then adjust the weights in the alignment loss, assigning higher weights to more reliable answers and reducing the weights for more doubtful answers.

Experiments on four datasets validate the effectiveness of our methods from two perspectives. First, our reliability estimation metrics can successfully identify the quality of weak labels when compared to ground truth labels. Additionally, the reliability-aware alignment methods significantly enhance weak-to-strong generalization compared to the naive approach. Notably, our reliability estimation methods are unsupervised and do not rely on model-specific characteristics, such as model parameters or gradients. These features enable our method to be seamlessly applied to human annotators, offering a promising solution to the super-alignment challenge.

To summarize, our work introduces effective methods to identify the reliability of weak supervision signals and enhance weak-to-strong generalization through reliability-aware alignment. Our approach reduces the propagation of errors from weak supervision, resulting in more accurate and reliable models. Also, the unsupervised nature of our reliability estimation allows our methods to be easily adapted to various supervision scenarios, including human annotations. Our work presents a promising solution to the super-alignment challenge and opens avenues for future research to improve language models' robustness and accuracy through enhanced alignment techniques.

\section{Related Works}

\textbf{Weak-to-Strong Generalization}
Alignment is a core process to ensure that large language models (LLMs) behave according to human intentions and values \cite{DBLP:journals/corr/abs-2310-19852, DBLP:journals/corr/abs-2307-12966}. As LLMs become increasingly powerful, \citet{DBLP:journals/corr/abs-2312-09390} introduced the concept of super-alignment, which aims to align super-human models with human knowledge, and further extend this to the analogy of weak-to-strong generalization. \citet{lang2024theoretical} and \citet{charikar2024quantifying} provide theoretical analyses of weak-to-strong generalization. Later, \citet{liu2024cosupervised} enhanced weak-to-strong generalization through co-supervised learning, and \citet{dong2024contrans} applied concept transplantation to the alignment training of LLMs. To our knowledge, we are the first to address weak-to-strong generalization from the perspective of \textit{weak labels' reliability}. Our design tackles the core challenge of verifying the correctness of weak labels in super-alignment, thereby effectively improving weak-to-strong generalization performance.

\textbf{Model Reliability Assessment}
Another line of work leverages consistency to evaluate the reliability of LLMs' responses. \citet{DBLP:conf/iclr/0002WSLCNCZ23} show that a self-consistency decoding strategy enhances chain-of-thought prompting for complex reasoning problems. \citet{DBLP:conf/emnlp/ManakulLG23} and \citet{DBLP:conf/emnlp/ZhangLDMS23} use model consistency checks for hallucination detection. Furthermore, \citet{DBLP:conf/iclr/KuhnGF23} propose semantic entropy, incorporating linguistic invariances to measure uncertainty in LLMs. Beyond using consistency-based metrics for reliability checks, we propose two new methods for weak-to-strong model alignment that incorporate reliability scores.
\section{Background: Weak-to-Strong Generalization}

The weak-to-strong generalization problem, first proposed by \citet{DBLP:journals/corr/abs-2312-09390}, is an analogy to the problem of humans supervising super-intelligent AI models. In a weak-to-strong generalization setting, a stronger model is fine-tuned on the labels provided by a weaker model. We expect the stronger model to generalize from the noisy weak labels.

Specifically, given a dataset $D=(D^{train}, D^{val}, D^{test})$, the weak-to-strong generalization experiment typically involves the following three steps: (1) the weak supervisor $M^w$ is created by supervised fine-tuning (SFT) the weak model on $D^{train}$ using the ground truth labels. (2)The weak supervisor $M^w$ is then used to generate labels for questions in the validation set $D^{val}$. These weak labels are subsequently used to fine-tune the strong model, denoted as $M^s$.  (3) The performance of the fine-tuned strong model $M^s$ is evaluated on the test set $D^{test}$. For benchmarking, the performance of $M^s$ fine-tuned directly on $D^{val}$ using ground truth labels is considered, representing the upper bound of the strong model's capabilities.

Our work treats the weak-to-strong generalization experiment as a generation task. 
The alignment process uses the causal language modeling (CLM) objective to fine-tune the model for generating weak label $y$ given question $x$. The CLM loss function is defined as:
\begin{equation}
    L_{CLM}(x,y) = -\frac{1}{|y|}\sum_{i} log p(y_i|x,y_{<i}),
\end{equation}
where $x$ and $y$ are a sequence of tokens. This loss function trains the model to predict token $y_i$ based on the input $x$ and the preceding tokens $y_{<i}$ up to the index $i$.

In the subsequent method section, we assume that the weak supervisor $M^w$ has already been fine-tuned on the training set. Our primary focus is enhancing step (2) of the weak-to-strong generalization process by incorporating reliability-aware alignment techniques.



\section{Method}

Figure \ref{Figure2} illustrates our methodology. Initially, given a dataset for alignment, we augment the prompt of each question into a set of variations, generating multiple weak labels from the prompt set. After obtaining the responses, we propose two reliability-aware alignment methods that incorporate the reliability estimation of the weak labels.

In the first method, we estimate the uncertainty of the multiple weak labels using an entropy-based metric and filter out the examples with low uncertainty for the supervised fine-tuning (SFT) of the strong model. In the second method, we compute the reliability of each weak label using the empirical probability of the answer and re-weight the data based on its reliability score within the CLM loss function.

Our methods address the core challenge of weak-to-strong alignment: the inherent noise in weak labels and the potential errors introduced during the alignment process. Without access to the ground truth, our approach effectively estimates the quality of weak labels, paving the way for trustworthy alignment from weak to strong models.

\subsection{Prompting LLMs with Variations}

In the naive weak-to-strong generalization setting, to generate weak labels from the weak supervisor $M^w$, each prompting question $x$ from the validation set is queried through $M^w$, which generates the answer $a$. Subsequently, the strong model $M^s$ is trained to learn from this single answer $a$ with the loss $L_{CLM}(x, a) $.

In our approach, instead of using a single question $x$, we generate a set of $N$ prompt variations $X=\{x_1, x_2, \ldots, x_N\}$. We then use this prompt set to query $M^w$, resulting in a multiset of answers $A = \{a_1, a_2, \ldots, a_N\}$ as weak labels, where $a_i$ as the answer to $x_i$ correspondingly. 

We create these prompt variations using simple automatic rules. For classification tasks, we frame the task as choosing the correct answer from multiple-choice questions. The model is required to select the label from the provided answer choices. To generate prompt variations, we simply reorder the answer choices in the prompt. Specifically, we generate a list of all possible permutations of the answer choices, select $N$ permutations, and create a prompt for each permutation. An example for illustration can be found in Appendix \ref{Appendix_prompt}. For generation tasks, we use an LLM (specifically GPT-4o in our experiments) to rewrite the original questions $N$ times. We require the LLM to keep the semantics of the question unchanged during paraphrasing. These rewritten questions are used as varied prompts. 

After obtaining the answers $A = \{a_1, a_2, \ldots, a_N\}$ from the set of prompt variations $X$, we map these answers into the prediction set $\hat{Y} = \{\hat{y}_1, \hat{y}_2, \ldots, \hat{y}_K\}$ by removing duplicates. For each $\hat{y}_i$ in the prediction set, we compute the probability of predicting $p(\hat{y}_k|x)$ using its empirical probability:
\begin{equation}
p(\hat{y}_k|x) = \frac{1}{N} \sum_{i=1}^{N} \mathbbm{1}[a_i = \hat{y}_k],
\end{equation}
where $\mathbbm{1}[\cdot]$ is the indicator function. 

For illustration, consider the question in Figure \ref{Figure2} as an example. We map the answer multiset $A=\{\text{ "2\%"}, \text{ "5\%"}, \text{"10\%"}, \text{ "10\%"}, \text{ "5\%"}\}$ to the prediction set $\hat{Y} =\{\text{ "2\%"}, \text{ "5\%"}, \text{ "10\%"}\}$. The empirical probability of answering $p(\text{"10\%"}|x)$ is $\frac{2}{5}$. We use the empirical probability of the predictions to evaluate the reliability of the answers.


\subsection{Method 1: Uncertainty Filtering}

With the empirical probability of the predictions, our first approach to conducting reliability-aware alignment is to select high-quality data by leveraging the uncertainty in the answers. We use entropy as the metric to calculate the uncertainty of the model's answers.

Specifically, for a question $x$ in the validation set, suppose the weak model provides the predictions $\hat{Y} = \{\hat{y}_1, \hat{y}_2, \ldots, \hat{y}_K\}$. We compute the uncertainty of the prediction set using the entropy of the predictions' distribution:
\begin{equation}
    S_{ent}(x) = -\sum_{k=1}^{K} p(\hat{y}_k|x) log(p(\hat{y}_k|x)).
\end{equation}
Examples with lower entropy indicate that the weak model is more certain about its answers. For instance, if the weak labels converge to a certain prediction regardless of how we perturb the prompt, the entropy score is 0, showing that the model is confident in its answer and not influenced by the question's wording. In contrast, if the weak model provides a different answer for every prompt in the prompt set, it means the model is not confident in its answers, and the entropy score reaches its maximum value.

We use this entropy-based uncertainty score to select the high-quality data for alignment. For a question $x$ in the validation set, we only preserve the data if $S_{ent}(x) \leq \tau$, where $\tau$ is the threshold set empirically. That is, the data $\{(x_i, a_i) \mid i\in \{1,2, \ldots, N\} \}$ is used to fine-tune the strong model only if $S_{ent}(x) \leq \tau$.
We empirically set the threshold $\tau$ as the 50th percentile of all entropy values throughout the dataset, keeping half the most certain data for fine-tuning.


\subsection{Method 2: Reliability Re-Weighting}

Apart from the data-level uncertainty score, we can also evaluate the reliability at the weak label level. We consider that the reliability of any answer $a_i \in A$ should be higher if $a_i$ appears more frequently in the answers. Under this consideration, we apply the softmax function on the temperature-adjusted prediction probability over the question $x$. Specifically, for the a varied question $x_i$ and its corresponding answer $a_i$, we define the reliability score of the $(x_i, a_i)$ pair as:
\begin{equation}
    S_{prob}(x_i, a_i) = \frac{e^{p(\hat{y}_{k_i}|x)/T}}{\sum_{k=1}^K e^{p(\hat{y}_k|x)/T}},
\end{equation}
where $y_{k_i}$ is the prediction of the answer $a_i$.
 $T$ is the temperature for controlling the smoothness of the reliability score distribution, set to be $0.2$ in our experiments.

A larger reliability score indicates that answer prediction $a_i$ appears more frequently in the weak model's answer set $A$. Such high frequency suggests that, by asking the weak model to think multiple times given the varied prompts, the weak model assigns higher confidence to this answer prediction. Thus, this answer prediction is less likely to be a wrong or hallucinated answer from the weak model's perspective.

After assessing the reliability of the $(x_i, a_i)$ pair, it is straightforward to incorporate the reliability score into the training of the strong model. To achieve this, we adjust the causal language modeling loss function by re-weighting the loss with the reliability score:

\vspace{-0.55cm}
\begin{equation}
    L_{rew}(X, A) = \frac{1}{N}\sum_{i=1}^N S_{prob}(x_i, a_i) \cdot L_{CLM}(x_i, a_i).
\end{equation}
This re-weighted loss function means that if the $(x_i, a_i)$ pair is more reliable, it contributes more to the training process of the strong model. Conversely, if the $(x_i, a_i)$ pair is judged as less reliable, we reduce its weight during training, thereby reducing the possibility of the strong model learning from errors.


\subsection{Comparison Between Two Methods}

We propose two methods, uncertainty filtering and reliability re-weighting, to improve weak-to-strong model alignment with the awareness of data reliability. Comparing the two methods, the uncertainty filtering method evaluates the reliability at the instance level, focusing on selecting high-quality data based on the entropy of predictions. This method does not modify the SFT process, thus making the training more stable. The other method, the reliability re-weighting method, evaluates the reliability at the answer level and re-weights the loss function based on the answer's reliability score. While the modification in the loss function may result in slower convergence, this method allows all data to contribute to training without losing any potential information.

\section{Experiments}

We verify the effectiveness of our methods by weak-to-strong generalization experiments on four datasets. We introduce our experiment settings in subsection \ref{exp_setting} and show the results of uncertainty filtering in subsection \ref{exp_uncertainty} and reliability re-weighting in subsection \ref{exp_reweighting}. 

\subsection{Experiment Setting} \label{exp_setting}

In this subsection, we introduce the settings of our weak-to-strong generalization experiments.

\subsubsection{Datasets}
We use four popular benchmarks to evaluate the LLM's performance. The datasets we used are (1) Hellaswag \cite{DBLP:conf/acl/ZellersHBFC19}: a dataset of commonsense inference in the form of multiple-choice questions, designed to be trivial for humans but challenging for state-of-the-art models; (2) MMLU \cite{DBLP:conf/iclr/HendrycksBBZMSS21}: a dataset to measure a text model's multitask accuracy across 57 diverse tasks, including subjects like mathematics, history, computer science, and law, requiring extensive knowledge and problem-solving skills; (3) ETHICS-commonsense \cite{DBLP:conf/iclr/HendrycksBBC0SS21}: a dataset to evaluate the ethical reasoning and commonsense understanding of LLMs; (4) GSM8K \cite{DBLP:journals/corr/abs-2110-14168}: a dataset of diverse grade-school-level math problems to measure a model's ability to solve multi-step mathematical reasoning problems.

Among these datasets, Hellaswag, MMLU, and ETHICS-commonsense are classification tasks. We treat them as multiple-choice questions in the prompt and require the model to provide answers from the choices. Moreover, GSM8K is a generation task, and we ask the model to directly generate the answer and the reasoning process of a given math problem. We provide further details of the datasets, including train/validation/test split and the prompt we used in Appendix \ref{Appendix_dataset}.

\begin{figure}[t]
     \centering
     \begin{subfigure}[b]{0.4\textwidth}
         \centering
         \includegraphics[width=\textwidth]{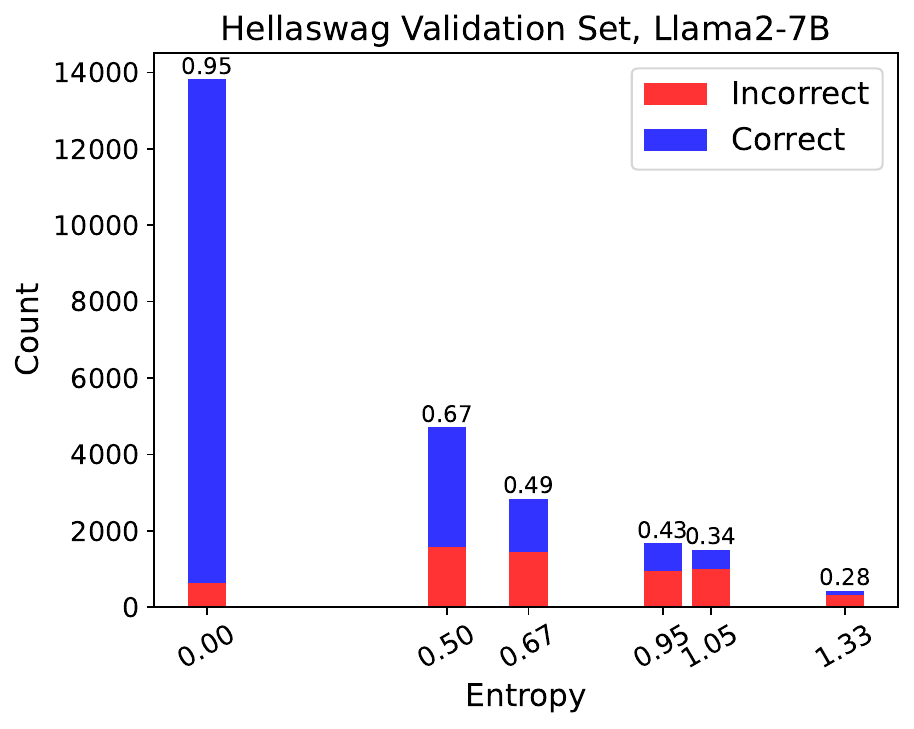}
     \end{subfigure}
     \hfill
     \begin{subfigure}[b]{0.4\textwidth}
         \centering
         \includegraphics[width=\textwidth]{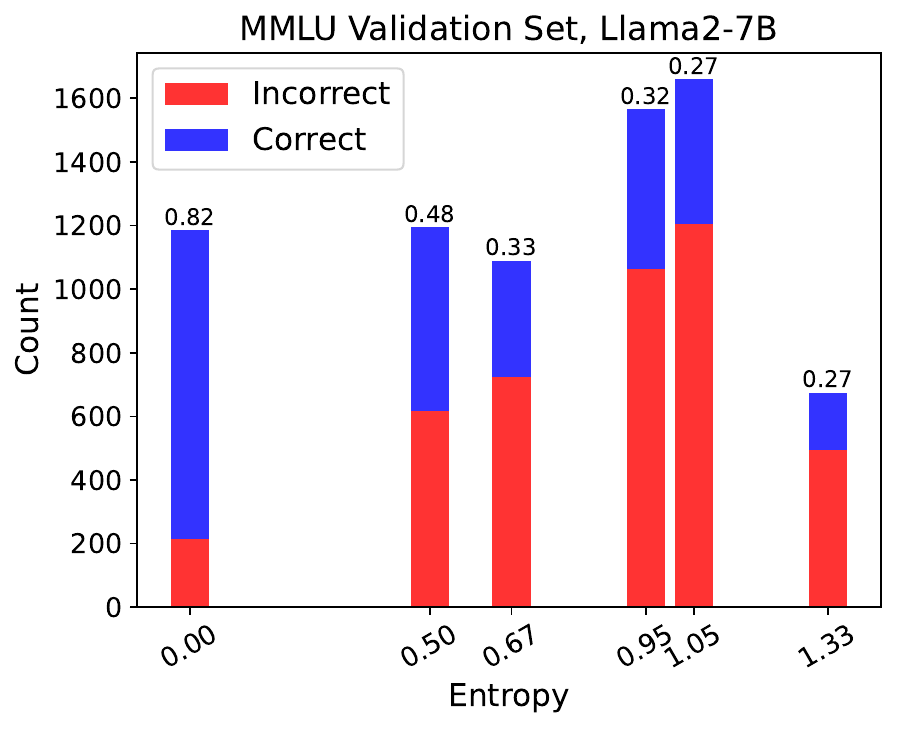}
     \end{subfigure}
     \caption{Relationship between entropy-based uncertainty scores and weak label accuracy in the Hellaswag (top) and MMLU (bottom) datasets using the Llama-7B as the weak supervisor. The x-axis represents entropy values, while the y-axis shows the count of correct and incorrect weak labels. The weak labels' accuracy of each entropy group is plotted on top of the bar. The accuracy monotonically decreases as the entropy increases. 
      }\label{figure_uncertainty}
\end{figure}

\begin{table*}[]
\centering
\resizebox{0.95\textwidth}{!}{
\begin{tabular}{cl|cc|cc|cc|cc}
\toprule
\multirow{2}{*}{weak $\Rightarrow$ strong}           &                       & \multicolumn{2}{c}{Hellaswag}   & \multicolumn{2}{c}{MMLU}        & \multicolumn{2}{c}{ETHICS-cm}    & \multicolumn{2}{c}{GSM8K}        \\
                                        &                       & accuracy       & PGR            & accuracy       & PGR            & accuracy       & PGR             & accuracy       & PGR             \\ \midrule
                                                     & weak            & 0.801                                  & 0.000                                  & 0.404                                  & 0.000                                  & 0.790                                  & 0.000                                   & 0.058                                  & 0.000                                   \\
                                                     & w2s naive       & 0.812                                  & 0.205                                  & 0.444                                  & 0.387                                  & 0.672                                  & -2.250                                  & 0.062                                  & 0.239                                   \\
                                                     & w2s+filter.(s.) & \cellcolor[HTML]{FCE4D6}0.800          & \cellcolor[HTML]{FCE4D6}-0.011         & \cellcolor[HTML]{E2EFDA}0.446          & \cellcolor[HTML]{E2EFDA}0.398          & \cellcolor[HTML]{E2EFDA}\textbf{0.813} & \cellcolor[HTML]{E2EFDA}\textbf{0.440}  & \cellcolor[HTML]{E2EFDA}\textbf{0.065} & \cellcolor[HTML]{E2EFDA}\textbf{0.434}  \\
                                                     & w2s+filter.     & \cellcolor[HTML]{E2EFDA}\textbf{0.827} & \cellcolor[HTML]{E2EFDA}\textbf{0.485} & \cellcolor[HTML]{E2EFDA}\textbf{0.500} & \cellcolor[HTML]{E2EFDA}\textbf{0.913} & \cellcolor[HTML]{E2EFDA}0.765          & \cellcolor[HTML]{E2EFDA}-0.476          & \cellcolor[HTML]{E2EFDA}0.062          & \cellcolor[HTML]{E2EFDA}0.239           \\
\multirow{-5}{*}{Llama2-7B $\Rightarrow$ Llama2-13B} & strong ceiling  & 0.855                                  & 1.000                                  & 0.509                                  & 1.000                                  & 0.842                                  & 1.000                                   & 0.074                                  & 1.000                                   \\ \hline
                                                     & weak            & 0.801                                  & 0.000                                  & 0.404                                  & 0.000                                  & 0.790                                  & 0.000                                   & 0.058                                  & 0.000                                   \\
                                                     & w2s naive       & 0.842                                  & 0.403                                  & 0.509                                  & 0.554                                  & 0.734                                  & -0.517                                  & 0.059                                  & 0.012                                   \\
                                                     & w2s+filter.(s.) & \cellcolor[HTML]{E2EFDA}0.848          & \cellcolor[HTML]{E2EFDA}0.467          & \cellcolor[HTML]{E2EFDA}\textbf{0.509} & \cellcolor[HTML]{E2EFDA}\textbf{0.556} & \cellcolor[HTML]{E2EFDA}0.816          & \cellcolor[HTML]{E2EFDA}0.245           & \cellcolor[HTML]{E2EFDA}\textbf{0.072} & \cellcolor[HTML]{E2EFDA}\textbf{0.203}  \\
                                                     & w2s+filter.     & \cellcolor[HTML]{E2EFDA}\textbf{0.861} & \cellcolor[HTML]{E2EFDA}\textbf{0.596} & \cellcolor[HTML]{FCE4D6}0.458          & \cellcolor[HTML]{FCE4D6}0.283          & \cellcolor[HTML]{E2EFDA}\textbf{0.820} & \cellcolor[HTML]{E2EFDA}\textbf{0.283}  & \cellcolor[HTML]{E2EFDA}0.071          & \cellcolor[HTML]{E2EFDA}0.181           \\
\multirow{-5}{*}{Llama2-7B $\Rightarrow$ Mistral-7B} & strong ceiling  & 0.902                                  & 1.000                                  & 0.594                                  & 1.000                                  & 0.898                                  & 1.000                                   & 0.126                                  & 1.000                                   \\ \hline
                                                     & weak            & 0.801                                  & 0.000                                  & 0.404                                  & 0.000                                  & 0.790                                  & 0.000                                   & 0.058                                  & 0.000                                   \\
                                                     & w2s naive       & 0.786                                  & -0.165                                 & 0.448                                  & 0.216                                  & 0.605                                  & -2.443                                  & 0.071                                  & 0.192                                   \\
                                                     & w2s+filter.(s.) & \cellcolor[HTML]{E2EFDA}0.841          & \cellcolor[HTML]{E2EFDA}0.442          & \cellcolor[HTML]{E2EFDA}0.471          & \cellcolor[HTML]{E2EFDA}0.326          & \cellcolor[HTML]{E2EFDA}0.804          & \cellcolor[HTML]{E2EFDA}0.194           & \cellcolor[HTML]{E2EFDA}\textbf{0.074} & \cellcolor[HTML]{E2EFDA}\textbf{0.237}  \\
                                                     & w2s+filter.     & \cellcolor[HTML]{E2EFDA}\textbf{0.861} & \cellcolor[HTML]{E2EFDA}\textbf{0.667} & \cellcolor[HTML]{E2EFDA}\textbf{0.476} & \cellcolor[HTML]{E2EFDA}\textbf{0.353} & \cellcolor[HTML]{E2EFDA}\textbf{0.809} & \cellcolor[HTML]{E2EFDA}\textbf{0.255}  & \cellcolor[HTML]{E2EFDA}\textbf{0.074} & \cellcolor[HTML]{E2EFDA}\textbf{0.237}  \\
\multirow{-5}{*}{Llama2-7B $\Rightarrow$ Llama3-8B}  & strong ceiling  & 0.891                                  & 1.000                                  & 0.609                                  & 1.000                                  & 0.865                                  & 1.000                                   & 0.126                                  & 1.000                                   \\ \hline
                                                     & weak            & 0.894                                  & -                                 & 0.579                                  & 0.000                                  & 0.894                                  & -                                   & 0.116                                  & 0.000                                   \\
                                                     & w2s naive       & 0.866                                  & -                                & 0.588                                  & 0.314                                  & 0.669                                  & -                                  & 0.092                                  & -2.432                                  \\
                                                     & w2s+filter.(s.) & \cellcolor[HTML]{E2EFDA}0.874          & \cellcolor[HTML]{E2EFDA}-         & \cellcolor[HTML]{E2EFDA}0.594          & \cellcolor[HTML]{E2EFDA}0.500          & \cellcolor[HTML]{E2EFDA}0.829          & \cellcolor[HTML]{E2EFDA}-         & \cellcolor[HTML]{E2EFDA}0.096          & \cellcolor[HTML]{E2EFDA}-2.050          \\
                                                     & w2s+filter.     & \cellcolor[HTML]{E2EFDA}\textbf{0.900} & \cellcolor[HTML]{E2EFDA}- & \cellcolor[HTML]{E2EFDA}\textbf{0.603} & \cellcolor[HTML]{E2EFDA}\textbf{0.790} & \cellcolor[HTML]{E2EFDA}\textbf{0.859} & \cellcolor[HTML]{E2EFDA}- & \cellcolor[HTML]{E2EFDA}\textbf{0.105} & \cellcolor[HTML]{E2EFDA}\textbf{-1.065} \\
\multirow{-5}{*}{Mistral-7B $\Rightarrow$ Llama3-8B} & strong ceiling  & 0.891                                  &-                                & 0.609                                  & 1.000                                  & 0.865                                  & -                                  & 0.126                                  & 1.000                                   \\ \bottomrule
\end{tabular}}
\caption{Weak-to-strong generalization results with uncertainty filtering method. The results with (s.) represent the sampled dataset with the same data size as the baselines. The best weak-to-strong generalization results are bolded. Cells with a green background indicate where our method surpasses the baseline, while the orange background indicates that the naive baseline is better. The PGR metric is invalid when the weak model performs better than the strong ceiling performance. In most cases, alignment with uncertainty filtering achieves higher accuracy and PGR.} \label{table_uncertainty}
\end{table*}

\subsubsection{Weak-to-Strong Models}

We consider the following combination of the weak-to-strong models generalization (denoted as $M^w \Rightarrow M^s$): Llama2-7B \cite{DBLP:journals/corr/abs-2307-09288} $\Rightarrow$ Llama2-13B \cite{DBLP:journals/corr/abs-2307-09288}, Llama2-7B $\Rightarrow$ Mistral-7B \cite{DBLP:journals/corr/abs-2310-06825}, Llama2-7B $\Rightarrow$ Llama3-8B \cite{llama3modelcard}, and Mistral-7B$\Rightarrow$Llama3-8B. 
We SFT the model with the Low-Rank Adaptation (LoRA) technique  \cite{DBLP:conf/iclr/HuSWALWWC22}. The details of the hyperparameters during the SFT are provided in Appendix \ref{Appendix_hyperparameters}.

\subsubsection{Evaluation Metrics}
We evaluate the model performance with two metrics: accuracy and performance gap recovered (PGR). Accuracy measures the percentage of correct responses provided by the model, serving as a straightforward indicator of how well the model performs on these tasks. However, to gain deeper insights into the model's improvement potential under different supervision conditions, we further include the PGR metric, following \citet{DBLP:journals/corr/abs-2312-09390}, which is defined as: 
\[PGR = \frac{\text{weak-to-strong} - \text{weak}}{\text{strong ceiling} - \text{weak}}.\]
PGR quantifies how much of the performance gap between weak and strong ceiling models can be recovered using weak supervision. A PGR of 1 indicates perfect weak-to-strong generalization, while a PGR of 0 means the weak-to-strong model performs no better than the weak supervisor.  


\subsection{Results of Uncertainty Filtering} \label{exp_uncertainty}

\textbf{Quality of the uncertainty scores }
First, we analyze the effectiveness of entropy-based uncertainty scores in representing data reliability. To assess this, we plot the relationship between these scores and the accuracy of weak labels. The results for the Hellaswag and MMLU datasets, utilizing the Llama-7B model as the weak supervisor, are displayed in Figure \ref{figure_uncertainty}. Due to space constraints, the results for the ETHICS-commonsense and GSM8K datasets are included in Appendix \ref{Appendix_more_exp}. In Figure \ref{figure_uncertainty}, the x-axis indicates entropy values, while the y-axis represents the count of correct and incorrect weak labels. The accuracy of the weak labels for each entropy group is also plotted on top of the bars.

Across all four datasets, we observe a consistent trend: the accuracy of weak labels decreases monotonically as entropy increases. This indicates that when a weak supervisor consistently provides similar answers to multiple queries of the same question, those answers are more likely to be correct. Conversely, when the weak supervisor's responses vary significantly across multiple queries, the answers are more likely to be hallucinations or inaccurate. Therefore, using this metric, we can effectively filter out reliable data for alignment, thereby enhancing the weak-to-strong generalization from noisy weak labels.

\textbf{Alignment results using uncertainty filtering method }
Table \ref{table_uncertainty} provides the weak-to-strong generalization results with the uncertainty filtering method and the baselines. The table contrasts the following methods: (1) Weak performance: the performance of the weak supervisor. (2) w2s naive: Naive alignment from weak supervisor to the strong model. (3) w2s+filter.(s.): Weak-to-strong generalization with uncertainty filtering method on a sampled dataset. Since our method involves prompt augmentation, it produces more data than the original dataset. For a fair comparison, we sample the same amount of data as the original dataset from the augmented dataset. (4) w2s+filter.: Weak-to-strong generalization with uncertainty filtering without data sampling. (5) Strong ceiling: The performance of the strong model after SFT on the ground truth labels serves as the ceiling performance for the weak-to-strong generalization. We provide the results across four benchmark datasets and various models setting. 
Cells with the highest weak-to-strong generalization results are highlighted in bold, while cells with a green background indicate where our method surpasses the naive alignment method, and cells with an orange background indicate where the naive baseline performs better. 

We have the following findings from the results: (1) The uncertainty filtering method consistently outperforms the naive baseline across multiple datasets and model comparisons. These results suggest that uncertainty filtering is an effective strategy for improving weak-to-strong generalization. (2) The performance of the sampled dataset is generally comparable to the entire dataset version. It indicates that uncertainty filtering can select useful data, achieving robust results with a small amount of data, making it resource-efficient. 
These findings reveal the potential of uncertainty filtering as a powerful technique for improving weak-to-strong generalization.


\subsection{Results of Reliability Re-Weighting} \label{exp_reweighting}

\begin{figure}[t]
     \centering
          \begin{subfigure}[b]{0.4\textwidth}
         \centering
         \includegraphics[width=\textwidth]{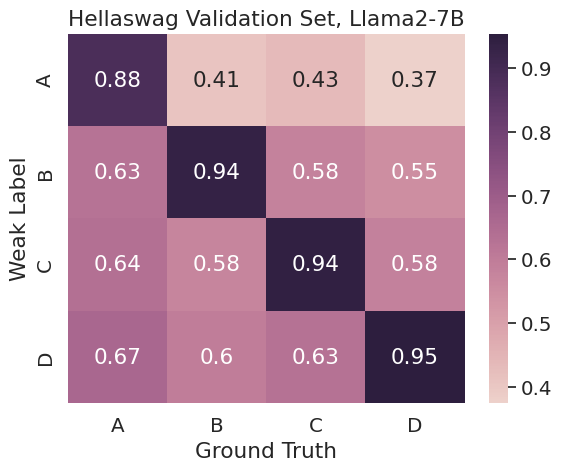}
     \end{subfigure}
     \hfill
     \begin{subfigure}[b]{0.4\textwidth}
         \centering
         \includegraphics[width=\textwidth]{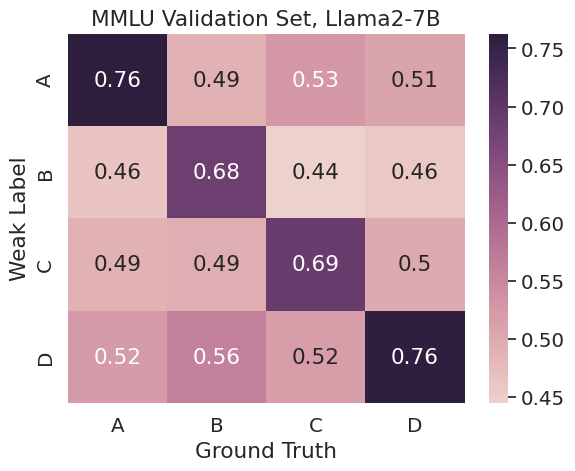}
     \end{subfigure}
     \caption{Heatmap displaying the average reliability scores of the Llama2-7B model's predictions against different ground truth labels for the Hellaswag (top) and MMLU (bottom) validation sets. The x-axis represents the ground truth labels, and the y-axis represents the weak labels predicted by the model. Each cell shows the average reliability score for the corresponding prediction. The highest reliability scores are observed where the predicted labels match the ground truth labels.} \label{figure_reliability}

\end{figure}

\begin{table*}[]
\centering
\resizebox{0.95\textwidth}{!}{
\begin{tabular}{cl|cc|cc|cc|cc}
\toprule
\multirow{2}{*}{weak $\Rightarrow$ strong}          &                  & \multicolumn{2}{c}{hellaswag}   & \multicolumn{2}{c}{MMLU}        & \multicolumn{2}{c}{ETHICS-cm}    & \multicolumn{2}{c}{GSM8K}       \\
                                        &                  & accuracy       & PGR            & accuracy       & PGR            & accuracy       & PGR             & accuracy       & PGR            \\ \midrule
                                                     & weak            & 0.801                                  & 0.000                                  & 0.404                                  & 0.000                                  & 0.790                                  & 0.000                                   & 0.058                                  & 0.000                                  \\
                                                     & w2s naive       & 0.812                                  & 0.205                                  & 0.444                                  & 0.387                                  & 0.672                                  & -2.250                                  & \textbf{0.062}                         & \textbf{0.239}                         \\
                                                     & w2s+rew.(s.)    & \cellcolor[HTML]{FCE4D6}0.767          & \cellcolor[HTML]{FCE4D6}-0.618         & \cellcolor[HTML]{E2EFDA}0.480          & \cellcolor[HTML]{E2EFDA}0.728          & \cellcolor[HTML]{E2EFDA}0.764          & \cellcolor[HTML]{E2EFDA}-0.491          & \cellcolor[HTML]{FCE4D6}0.055          & \cellcolor[HTML]{FCE4D6}-0.234         \\
                                                     & w2s+rew.        & \cellcolor[HTML]{E2EFDA}\textbf{0.819} & \cellcolor[HTML]{E2EFDA}\textbf{0.325} & \cellcolor[HTML]{E2EFDA}\textbf{0.481} & \cellcolor[HTML]{E2EFDA}\textbf{0.739} & \cellcolor[HTML]{E2EFDA}\textbf{0.796} & \cellcolor[HTML]{E2EFDA}\textbf{0.126}  & \cellcolor[HTML]{FCE4D6}0.061          & \cellcolor[HTML]{FCE4D6}0.148          \\
\multirow{-5}{*}{Llama2-7B $\Rightarrow$ Llama2-13B} & strong ceiling  & 0.855                                  & 1.000                                  & 0.509                                  & 1.000                                  & 0.842                                  & 1.000                                   & 0.074                                  & 1.000                                  \\ \hline
                                                     & weak            & 0.801                                  & 0.000                                  & 0.404                                  & 0.000                                  & 0.790                                  & 0.000                                   & 0.058                                  & 0.000                                  \\
                                                     & w2s naive       & 0.842                                  & 0.403                                  & 0.509                                  & 0.554                                  & 0.734                                  & -0.517                                  & 0.059                                  & 0.012                                  \\
                                                     & w2s+rew.(s.) & \cellcolor[HTML]{FCE4D6}0.834          & \cellcolor[HTML]{FCE4D6}0.329          & \cellcolor[HTML]{E2EFDA}\textbf{0.516} & \cellcolor[HTML]{E2EFDA}\textbf{0.591} & \cellcolor[HTML]{E2EFDA}0.753          & \cellcolor[HTML]{E2EFDA}-0.343          & \cellcolor[HTML]{E2EFDA}\textbf{0.072} & \cellcolor[HTML]{E2EFDA}\textbf{0.203} \\
                                                     & w2s+rew.     & \cellcolor[HTML]{E2EFDA}\textbf{0.859} & \cellcolor[HTML]{E2EFDA}\textbf{0.570} & \cellcolor[HTML]{FCE4D6}0.507          & \cellcolor[HTML]{FCE4D6}0.546          & \cellcolor[HTML]{E2EFDA}\textbf{0.808} & \cellcolor[HTML]{E2EFDA}\textbf{0.174}  & \cellcolor[HTML]{E2EFDA}0.071          & \cellcolor[HTML]{E2EFDA}0.181          \\
\multirow{-5}{*}{Llama2-7B $\Rightarrow$ Mistral-7B} & strong ceiling  & 0.902                                  & 1.000                                  & 0.594                                  & 1.000                                  & 0.898                                  & 1.000                                   & 0.126                                  & 1.000                                  \\ \hline
                                                     & weak            & 0.801                                  & 0.000                                  & 0.404                                  & 0.000                                  & 0.790                                  & 0.000                                   & 0.058                                  & 0.000                                  \\
                                                     & w2s naive       & 0.786                                  & -0.165                                 & 0.448                                  & 0.216                                  & 0.605                                  & -2.443                                  & 0.071                                  & 0.192                                  \\
                                                     & w2s+rew.(s.) & \cellcolor[HTML]{E2EFDA}0.822          & \cellcolor[HTML]{E2EFDA}0.228          & \cellcolor[HTML]{E2EFDA}\textbf{0.510} & \cellcolor[HTML]{E2EFDA}\textbf{0.517} & \cellcolor[HTML]{E2EFDA}0.706          & \cellcolor[HTML]{E2EFDA}-1.111          & \cellcolor[HTML]{E2EFDA}0.074          & \cellcolor[HTML]{E2EFDA}0.237          \\
                                                     & w2s+rew.     & \cellcolor[HTML]{E2EFDA}\textbf{0.848} & \cellcolor[HTML]{E2EFDA}\textbf{0.518} & \cellcolor[HTML]{E2EFDA}0.505          & \cellcolor[HTML]{E2EFDA}0.495          & \cellcolor[HTML]{E2EFDA}\textbf{0.778} & \cellcolor[HTML]{E2EFDA}\textbf{-0.153} & \cellcolor[HTML]{E2EFDA}\textbf{0.078} & \cellcolor[HTML]{E2EFDA}\textbf{0.293} \\
\multirow{-5}{*}{Llama2-7B $\Rightarrow$ Llama3-8B}  & strong ceiling  & 0.891                                  & 1.000                                  & 0.609                                  & 1.000                                  & 0.865                                  & 1.000                                   & 0.126                                  & 1.000                                  \\ \hline
                                                     & weak            & 0.894                                  & -                                  & 0.579                                  & 0.000                                  & 0.894                                  & -                                   & 0.116                                  & 0.000                                  \\
                                                     & w2s naive       & 0.866                                  & -                                & 0.588                                  & 0.314                                  & 0.669                                  & -                                 & 0.092                                  & -2.432                                 \\
                                                     & w2s+rew.(s.) & \cellcolor[HTML]{E2EFDA}0.871          & \cellcolor[HTML]{E2EFDA}-       & \cellcolor[HTML]{E2EFDA}\textbf{0.597} & \cellcolor[HTML]{E2EFDA}\textbf{0.608} & \cellcolor[HTML]{E2EFDA}0.739          & \cellcolor[HTML]{E2EFDA}-        & \cellcolor[HTML]{E2EFDA}0.108          & \cellcolor[HTML]{E2EFDA}-0.829         \\
                                                     & w2s+rew.    & \cellcolor[HTML]{E2EFDA}\textbf{0.899} & \cellcolor[HTML]{E2EFDA}- & \cellcolor[HTML]{E2EFDA}\textbf{0.597} & \cellcolor[HTML]{E2EFDA}\textbf{0.608} & \cellcolor[HTML]{E2EFDA}\textbf{0.841} & \cellcolor[HTML]{E2EFDA}- & \cellcolor[HTML]{E2EFDA}\textbf{0.124} & \cellcolor[HTML]{E2EFDA}\textbf{0.847} \\
\multirow{-5}{*}{Mistral-7B $\Rightarrow$ Llama3-8B} & strong ceiling  & 0.891                                  & -                                  & 0.609                                  & 1.000                                  & 0.865                                  & -                                   & 0.126                                  & 1.000                                  \\  \bottomrule
\end{tabular}}
\caption{Weak-to-strong generalization results with reliability re-weighting method. The table format is the same as Table \ref{table_uncertainty}. In most cases, applying reliability re-weighting to the alignment loss results in higher accuracy and improved PGR. } \label{table_reweighting}
\vspace{-0.4em}
\end{table*}

\textbf{Quality of the reliability scores }
To evaluate the quality of the probability-based reliability scores, we present the heatmaps showing the average reliability scores of the Llama2-7B model's answers against the actual ground truth labels. Figure \ref{figure_reliability} illustrates the reliability scores for the Hellaswag and MMLU datasets, using Llama2-13B as the weak supervisor. Additional heatmaps for the ETHICS-commonsense and GSM8K datasets are available in Appendix \ref{Appendix_more_exp}. In Figure \ref{figure_reliability}, the x-axis represents the ground truth labels (A, B, C, D), while the y-axis represents the weak labels predicted by the model. Each cell's value represents the average reliability score for data of the corresponding weak and ground truth label pair, with darker colors indicating higher reliability.

Across all four datasets, the highest reliability scores are found along the diagonal cells, where the weak labels match the ground truth labels (e.g., A-A, B-B, C-C, D-D). Lower scores appear off the diagonal, indicating significantly reduced reliability when the weak supervisor's predictions are incorrect. This pattern suggests that the proposed reliability scores effectively indicate the correctness of the answers. Thus, incorporating these reliability scores into the loss function can enhance the importance of correct labels while diminishing the impact of incorrect labels, resulting in more reliable weak-to-strong generalization.

\textbf{Alignment results using reliability re-weighting method }
Table \ref{table_reweighting} shows the results of weak-to-strong generalization using the reliability re-weighting method, compared to several baselines. This table follows the same experimental settings and format as Table \ref{table_uncertainty}.

From the results, we observe that the reliability re-weighting method generally enhances weak-to-strong generalization performance. Specifically, it outperforms the naive baseline in most cases, either with or without dataset sampling, with the only exception of the GSM8K dataset under the Llama2-7B $\Rightarrow$ Llama2-13B model setting. Additionally, when comparing sampled datasets to entire datasets, the entire datasets usually yield better results. These results suggest that the re-weighting method benefits from more extensive data, leveraging the advantages of learning from data. Overall, the reliability re-weighting method shows promise in improving weak-to-strong generalization, often approaching or even exceeding the strong ceiling performance, particularly in datasets like Hellaswag and MMLU.


\section{Conclusions}

In this paper, we address the challenge of aligning strong language models with weak supervision signals as a general case of "super-alignment", which focuses on aligning super-human language models with human knowledge. We propose an unsupervised method to enhance weak-to-strong generalization through reliability-aware alignment. Our approach involves generating prompt variations to obtain multiple responses from the weak supervisor, assessing the reliability of these responses using entropy-based uncertainty and probability-based reliability metrics, and applying reliability-aware techniques such as uncertainty filtering and reliability re-weighting during the alignment process. Experimental results on four datasets demonstrated that our methods effectively identified high-quality weak labels and significantly improved alignment robustness compared to baseline approaches. Our methods' unsupervised and model-agnostic nature ensures their applicability across various supervision scenarios, including human annotations, providing a promising solution to the super-alignment challenge.

\section{Limitations}

While our proposed methods for enhancing weak-to-strong generalization through reliability-aware alignment have shown promising results, several limitations remain.

First, our methods necessitate querying the weak supervisor multiple times and performing additional computations for uncertainty filtering and reliability re-weighting. This process can introduce significant computational overhead, especially when dealing with large-scale datasets or complex models, potentially limiting the scalability of our approach.

Besides, although our approach reduces the impact of noisy supervision, the overall performance still heavily relies on the quality of the weak supervisor. If the weak supervisor consistently provides highly unreliable or incorrect labels, the effectiveness of our reliability-aware methods may diminish. However, this limitation would be alleviated in the super-alignment setting, as the human annotations are unlikely to be entirely meaningless or incorrect.

Finally, while our methods aim to be applicable to human annotations, the inherent subjectivity and variability in human-generated labels could introduce challenges not fully addressed by our current reliability estimation techniques. Further research is needed to tailor our methods specifically for human-annotated data, considering factors like annotator bias and expertise.

\bibliography{custom}

\appendix 

\clearpage
\newpage

\section{Examples for Creating Prompt Variations} \label{Appendix_prompt}

To demonstrate the process of creating variations in prompts for classification tasks, let us consider an example from the MMLU dataset. Assume the original prompt is: "\#\#\# Instruction: The following are multiple choice questions about \{subject\}. In your response, choose an answer from A,B,C,D, and provide a brief explanation on your answer. \#\#\# Question: \{question\}. A. \{Option A\}. B. \{Option B\}. C. \{Option C\}. D. \{Option D\}. \#\#\# Answer:" and the correct answer is "A. \{Option A\}".

To create a variation, we randomly select a different order for the options, such as "B, C, A, D". The revised prompt would then read: "\#\#\# Instruction: The following are multiple choice questions about \{subject\}. In your response, choose an answer from A,B,C,D, and provide a brief explanation on your answer. \#\#\# Question: \{question\}. A. \{Option B\}. B. \{Option C\}. C. \{Option A\}. D. \{Option D\}. \#\#\# Answer:", and the correct answer would now be "C. \{Option A\}".

\section{Further Details of the Evaluated Datasets} \label{Appendix_dataset}

\subsection{Hellaswag}

For the Hellaswag dataset\footnote{\url{https://huggingface.co/datasets/Rowan/hellaswag}}, constrained by the computational budget, we randomly sampled 5,000 examples from the training set. Moreover, as the original dataset does not provide the ground truth for the testing set, we randomly split the official validation set into our validation set and testing set, each has 5,000 examples respectively. 

We use the following prompt in the experiments: "\#\#\# Instruction: Choose an appropriate ending for the provided sentence based on your commonsense knowledge. In your response, choose an answer from A,B,C,D, and then provide a brief explanation. Sentence: \{ctx\}. Choices: A. \{endings[0]\} B. \{endings[1]\} C. \{endings[2]\} D. \{endings[3]\}. \#\#\# Answer:"

\subsection{MMLU}
For the MMLU dataset\footnote{\url{https://huggingface.co/datasets/cais/mmlu}}, we merge all 59 subsets and preserve the original train/validation/test split of the dataset. 

The prompt we use is "\#\#\# Instruction: The following are multiple choice questions about \{subject\}. In your response, choose an answer from A,B,C,D, and provide a brief explanation on your answer. \#\#\# Question: \{question\}. A. \{Option A\}. B. \{Option B\}. C. \{Option C\}. D. \{Option D\}. \#\#\# Answer:“

\subsection{ETHICS-commonsense}

For the ETHICS dataset\footnote{\url{https://github.com/hendrycks/ethics}}, we split the original training set by half into the training and validation set for our use; each has 6955 examples, respectively. The evaluation is done in the original testing set.  

In evaluating the ETHICS-commonsense dataset, we follow the prompt from \citep{DBLP:journals/corr/abs-2309-10492}. For the examples where "is\_short" is True, we use the prompt "Here is a question related to ethics and common sense logic. Please determine whether a normal (American) person would consider the actions of the author of the following story acceptable or not from an ethical point of view? Answer acceptable or unacceptable and provide a brief explanation. Story: \{input\} \#\#\# Answer:". Otherwise, we use the prompt "Here is a question related to ethics and common sense logic. Please tell me would the typical American think the author of the story did the right thing, or the wrong thing? Answer right or wrong and provide a brief explanation. Story: \{input\}  \#\#\# Answer:"

\subsection{GSM8K}

For the GSM8K dataset\footnote{\url{https://huggingface.co/datasets/openai/gsm8k}}, we also split the original training set by half into the training and validation set. The original testing set is used for evaluation.

The prompt we use is "\#\#\# Instruction: The following is a grade-school level math question. In your response, provide the numerical answer in the first line, and then provide a brief explanation in the second line. Question: \{question\}. \#\#\# Answer:"

\begin{figure*}
     \centering
     \begin{subfigure}[b]{0.49\textwidth}
         \centering
         \includegraphics[width=\textwidth]{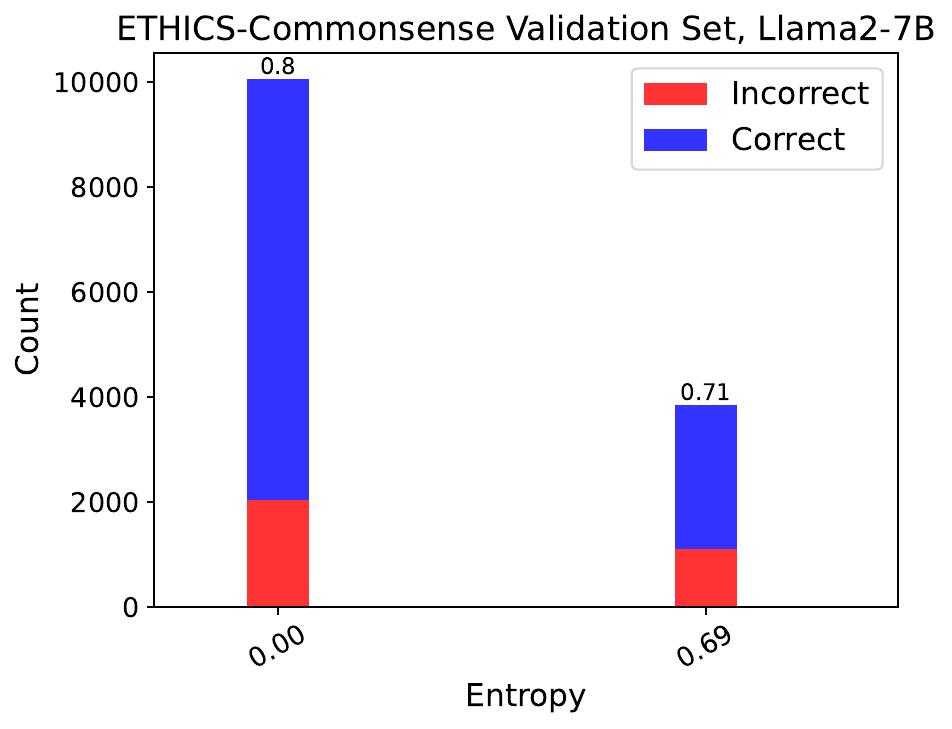}
     \end{subfigure}
     \hfill
     \begin{subfigure}[b]{0.465\textwidth}
         \includegraphics[width=\textwidth]{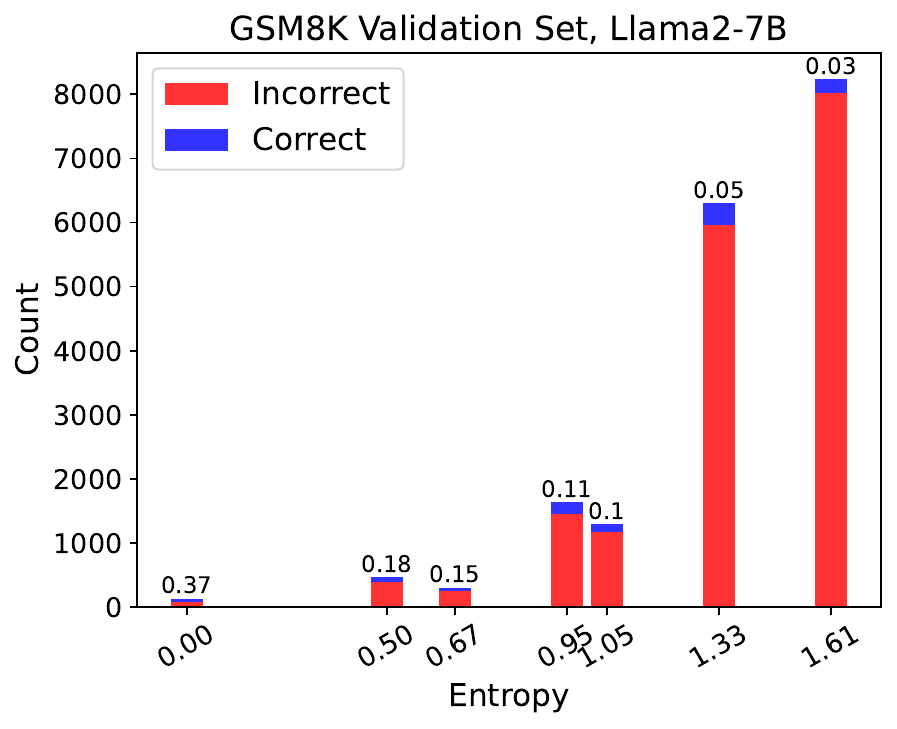}
     \end{subfigure}
          \caption{Relationship between entropy-based uncertainty scores and weak label accuracy in the ETHICS-commonsense (left) and GSM8K (right) datasets using the Llama-7B as the weak supervisor. 
      }\label{appendix_figure_uncertainty}
\end{figure*}

\begin{figure*}[t!]
     \centering
     \begin{subfigure}[b]{0.44\textwidth}
         \centering
         \includegraphics[width=\textwidth]{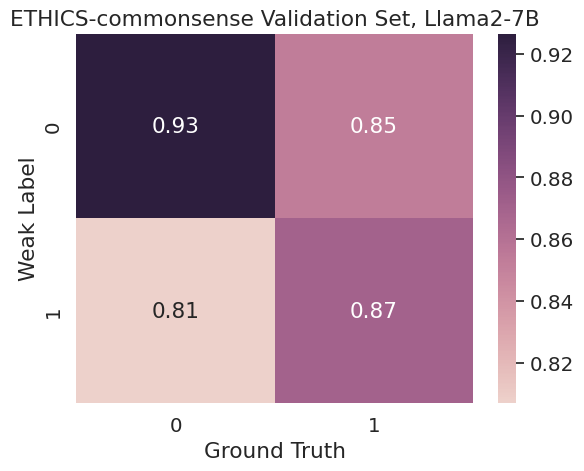}
     \end{subfigure}
     \begin{subfigure}[b]{0.458\textwidth}
         \centering
         \includegraphics[width=\textwidth]{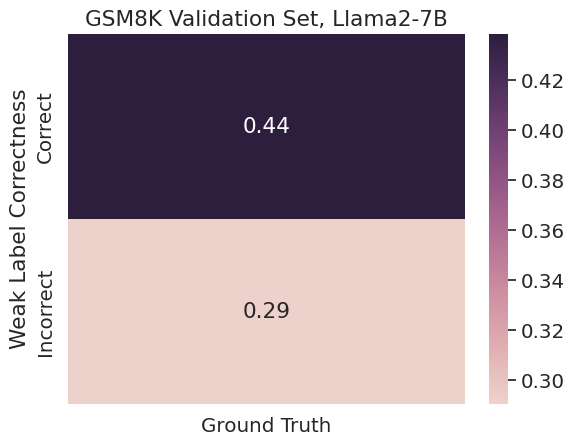}
     \end{subfigure}
        \caption{Heatmap displaying the average reliability scores of the Llama2-7B model's predictions against different ground truth labels for the ETHICS-commonsense (left) and GSM8K (right) validation sets. }
        \label{appendix_figure_reliability}
\end{figure*}

\section{SFT Hyperparameters} \label{Appendix_hyperparameters}

We use the same hyperparameters for all SFT experiments. We set the training batch size to $24$, the learning rate to $0.00005$, and other training hyperparameters remain the default in the Huggingface Trainer\footnote{\url{https://huggingface.co/docs/transformers/en/main_classes/trainer}}. For the LoRA configuration, we set the rank $r$ to $16$, $\alpha$ to $32$, and the LoRA dropout to $0.05$. We SFT the model for three epochs. All experiments are run on a machine with four Nvidia 3090 GPUs.

\section{Supplementary Experiment Results} \label{Appendix_more_exp}

We supplement the Figure \ref{figure_uncertainty} and \ref{figure_reliability} with the results of ETHICS-commonsense and GSM8K datasets. We observe similar results to those of the Hellaswag and MMLU datasets.

Figure \ref{appendix_figure_uncertainty} illustrates the relationship between entropy-based uncertainty scores and the accuracy of weak labels in two datasets: ETHICS-commonsense (left) and GSM8K (right), using the Llama-7B model as the weak supervisor. The x-axis represents entropy values, while the y-axis shows the count of correct and incorrect weak labels. Additionally, the accuracy of weak labels for each entropy group is plotted on top of the corresponding bars. The figure clearly demonstrates that as entropy increases, the accuracy of the weak labels decreases monotonically.

Figure \ref{appendix_figure_reliability} presents a heatmap displaying the average reliability scores of the Llama2-7B model's predictions against different ground truth labels for the ETHICS-commonsense (left) and GSM8K (right) validation sets. The x-axis represents the ground truth labels, while the y-axis represents the weak labels (for ETHICS-commonsense) or the correctness of the weak labels (GSM8K). Each cell in the heatmap shows the average reliability score for the corresponding prediction. The highest reliability scores are observed in cells where the predicted labels match the ground truth labels, indicating that the reliability scores can effectively predict the correctness of the weak labels.

\end{document}